\documentclass{article}

\usepackage{arxiv}

\usepackage[utf8]{inputenc} 
\usepackage[T1]{fontenc}    
\usepackage{hyperref}       
\usepackage{url}            
\usepackage{booktabs}       
\usepackage{amsfonts}       
\usepackage{nicefrac}       
\usepackage{microtype}      
\usepackage{graphicx}
\usepackage{lipsum}
\usepackage{listings}
\usepackage{color}
\usepackage{longtable}
\usepackage{multirow}
\usepackage{amsmath} 
\usepackage[svgnames]{xcolor}
\usepackage{hyperref}
\hypersetup{
    colorlinks=true,
    linkcolor=Blue,
    citecolor=Green,
    urlcolor=Navy,
}
\usepackage{pgfplots} 
\pgfplotsset{compat=1.18}
\usetikzlibrary{patterns.meta} 

\graphicspath{ {./images/} }

\title{DSCC-HS: A Dynamic Self-Reinforcing Framework for Hallucination Suppression in Large Language Models}

\author{
 Xiao Zheng \\
 School of Computing and Technology\\
 China University of Petroleum\\
 Qingdao \\
 \texttt{BY2007040229@s.upc.edu} \\
}

\begin{document}
\maketitle

\begin{abstract}
Hallucination remains a critical barrier to the reliable deployment of Large Language Models (LLMs) in high-stakes applications. Existing mitigation strategies, such as Retrieval-Augmented Generation (RAG) and post-hoc verification, are often reactive, inefficient, or fail to address the root cause within the generative process. Inspired by dual-process cognitive theory, we propose \textbf{D}ynamic \textbf{S}elf-reinforcing \textbf{C}alibration for \textbf{H}allucination \textbf{S}uppression (DSCC-HS), a novel, proactive framework that intervenes directly during autoregressive decoding. DSCC-HS operates via a two-phase mechanism: (1) During training, a compact proxy model is iteratively aligned into two adversarial roles---a Factual Alignment Proxy (FAP) and a Hallucination Detection Proxy (HDP)---through contrastive logit-space optimization using augmented data and parameter-efficient LoRA adaptation. (2) During inference, these frozen proxies dynamically steer a large target model by injecting a real-time, vocabulary-aligned steering vector (computed as the difference between FAP and HDP logits) at each decoding step, requiring no modification to the target model. This plug-and-play approach enables lightweight, scalable, and proactive hallucination suppression. Our experiments on the TruthfulQA and BioGEN benchmarks demonstrate DSCC-HS's state-of-the-art performance. On TruthfulQA, DSCC-HS achieves a \textbf{49.82\%} accuracy and a \textbf{99.2\%} Factual Consistency Rate (FCR) while reducing the hallucination score to \textbf{0.8}, significantly outperforming strong baselines like ITI and DOLA. On the long-form BioGEN benchmark, DSCC-HS attains the highest FActScore of \textbf{46.50} and the lowest incorrectness rate of \textbf{11.49}, showcasing its robustness in generating factually consistent long texts. These results validate DSCC-HS as a principled and efficient solution for enhancing LLM factuality.
\end{abstract}

\section{Introduction}

Large Language Models (LLMs) have ushered in a new era of artificial intelligence, showcasing unprecedented capabilities across a wide array of natural language tasks. Models such as GPT-4 \cite{achiam2023gpt} and LLaMA 2 \cite{touvron2023llama} have demonstrated remarkable "emergent abilities" \cite{wei2022emergent} as their scale increases, leading to their extensive application in dialogue systems \cite{zhang-etal-2023-sgp} and beyond \cite{liu2023summary}. However, despite these advancements, a persistent and critical challenge remains: hallucination \cite{ji2023survey}. This phenomenon, where LLMs generate seemingly plausible but factually incorrect or nonsensical content, fundamentally undermines their reliability and trustworthiness, particularly in high-stakes domains like medicine, finance, and law \cite{chang2024survey, liu2023trustworthy}. A growing body of research has been dedicated to understanding and cataloging this issue \cite{huang2025survey, zhang2025siren}.

Existing methods to mitigate hallucination largely fail due to inherent limitations in their design. The first paradigm, Retrieval-Augmented Generation (RAG) \cite{shi2023ralm}, attempts to ground models with external knowledge but often suffers from \textit{contextual ignorance}, where the model fails to correctly utilize or integrate the provided information, leading to factual inaccuracies \cite{tonmoy2024comprehensive}. The second paradigm, post-hoc verification, is fundamentally \textit{reactive and inefficient}. It attempts to correct errors after generation, which not only fails to address the root cause of the hallucination within the generative process but also introduces significant computational overhead. This leaves a critical need for a framework that can intervene proactively to ensure factual consistency from the outset.

Inspired by Nobel laureate Daniel Kahneman's dual-process cognitive theory \cite{kahneman}, which distinguishes between the fast, intuitive System 1 and the slow, deliberative System 2, we introduce the Dynamic Self-reinforcing Calibration for Hallucination Suppression (DSCC-HS). Inspired by Nobel laureate Daniel Kahneman's dual-process cognitive theory \cite{kahneman}, which distinguishes between the fast, intuitive System 1 and the slow, deliberative System 2, we introduce the Dynamic Self-reinforcing Calibration for Hallucination Suppression (DSCC-HS). Our framework establishes a novel two-phase architecture: first, during training, we iteratively align a compact proxy model into two adversarial roles—Factual Alignment Proxy (FAP) and Hallucination Detection Proxy (HDP)—via contrastive logit-space optimization, leveraging augmented data and parameter-efficient LoRA adaptation; second, during inference, we deploy these frozen proxies to dynamically steer the output of a large target model through a vocabulary-aligned steering vector, computed as the difference between FAP and HDP logits at each decoding step. This plug-and-play mechanism requires no modification to the target model’s parameters, enabling real-time, lightweight, and proactive hallucination suppression. By explicitly encoding the directionality of factuality versus hallucination in the proxy representation space, DSCC-HS intervenes directly in the generative process, offering a principled, efficient, and scalable solution to one of the most pressing challenges in modern LLM deployment.

To summarize, our contributions are listed as follows.

\begin{itemize}
    \item \textbf{A Novel Dual-Proxy Alignment Framework:} We propose the first method to explicitly train adversarial proxy models—Factual Alignment Proxy (FAP) and Hallucination Detection Proxy (HDP)—via iterative contrastive logit-space optimization, effectively carving out “factual” and “hallucinatory” manifolds in representation space without requiring human-labeled contrastive pairs at scale.
    
    \item \textbf{Parameter-Efficient and Iterative Training:} We introduce a lightweight, LoRA-based iterative alignment procedure that refines the FAP over multiple rounds while freezing the HDP after initialization, achieving strong factual specialization with minimal trainable parameters and no architectural modification to the base model.
    
    \item \textbf{Plug-and-Play Inference-Time Steering:} We design a non-invasive, real-time decoding mechanism that dynamically guides large target models by injecting a vocabulary-projected steering vector—computed as the difference between FAP and HDP logits—at each generation step, requiring zero retraining or fine-tuning of the target model.
    
    \item \textbf{Data-Augmented Proxy Training:} We enhance proxy generalization through a multi-strategy dataset augmentation pipeline—including question paraphrasing, answer perturbation, and external data supplementation—enabling robust alignment even with limited human-annotated factuality labels.
    
    \item \textbf{Proactive Hallucination Suppression:} Unlike reactive post-hoc methods, our framework intervenes directly within the autoregressive generation loop, suppressing hallucinations at their source by continuously nudging the target model toward factually consistent token distributions.
\end{itemize}

\section{Related Works}

\subsection{Hallucinations in Large Language Models}

Hallucinations in Large Language Models (LLMs) are a critical issue where generated content appears plausible but deviates from established facts \cite{chen2023factchd, li2023generative, zhang2025siren, tonmoy2024comprehensive}. We align with the perspective that an LLM's acquired knowledge should accurately mirror established world knowledge \cite{yang2024alignment}. In this study, we specifically focus on a type of ``unfaithful hallucination,'' where LLMs produce factually incorrect statements despite possessing relevant knowledge \cite{evans2021truthful, park2024ai, li2023inference}. Instead of broadly targeting the enhancement of LLMs' factuality \cite{sun2023aligning, zhou2023lima, lightman2023let, peng2023check, li2023chain, mallen2022not}, our goal is to align models to reliably convey accurate information when they have sufficient knowledge.

\subsection{Hallucination Mitigation Techniques}

Large language model (LLM) hallucinations pose a critical challenge in the advancement of artificial intelligence, prompting a wide array of research into \textbf{mitigation strategies}, which can be broadly categorized into three principal classes.

A prominent strategy within this realm is \textbf{post-hoc correction}, which involves the refinement of generated outputs after they have been produced. Techniques under this category, such as self-consistency mechanisms, aim to improve factual accuracy by evaluating the consistency across multiple responses from the model \cite{kadavath2022language, ren2023self, tian2023just, madaan2023self, dhuliawala2023chain, wang2022self}. Despite their proven effectiveness in certain contexts, the performance of these methods is inherently dependent on the model's internal capabilities, particularly its reasoning and generalization capacities.

Another widely explored strategy, \textbf{inference-time intervention}, seeks to influence the model's internal representations during the generation process to steer the output towards factual correctness. For instance, contrastive decoding techniques have been employed to explicitly encourage the model to generate more truthful responses \cite{li2023inference, chuang2023dola, li2022contrastive, zhang2023alleviating}. However, these methods are often constrained by their reliance on domain-specific knowledge or manually crafted rules, limiting their generalizability to a wider array of tasks.

ly, \textbf{alignment training} represents a direct approach to optimizing LLMs for factuality, achieved through supervised fine-tuning (SFT) on high-quality, annotated datasets \cite{wang2022self, zhang2022toward}, or through reinforcement learning from human feedback (RLHF) \cite{ouyang2022training}. While these methods have demonstrated considerable success in improving model accuracy, they require substantial computational resources and human labor, making them resource-intensive and challenging to scale.

\section{Methodology}
\label{sec:method}

Our proposed methodology, DSCC-HS, is structured into two sequential stages: \textit{Proxy Model Alignment}, a training-time procedure to distill factual knowledge and its antithesis into specialized proxy models, and \textit{Proxy-Guided Inference}, an inference-time mechanism to steer a large target model towards factuality. This integrated approach allows for the efficient control of large model outputs without modifying their internal parameters during inference.

\subsection{Phase 1: Iterative Alignment of Proxy Models}
\label{sec:phase1}

The first phase transforms a compact, general-purpose Large Language Model (LLM)---specifically, Llama-3.2-1B-Instruct---into a pair of adversarial proxy models. These proxies, designated the Factual Alignment Proxy (FAP) and the Hallucination Detection Proxy (HDP), are trained to represent opposing directions in the model's representation space corresponding to factual accuracy and hallucinatory content, respectively.

\begin{figure}[htbp]
    \centering
    \includegraphics[width=1\textwidth]{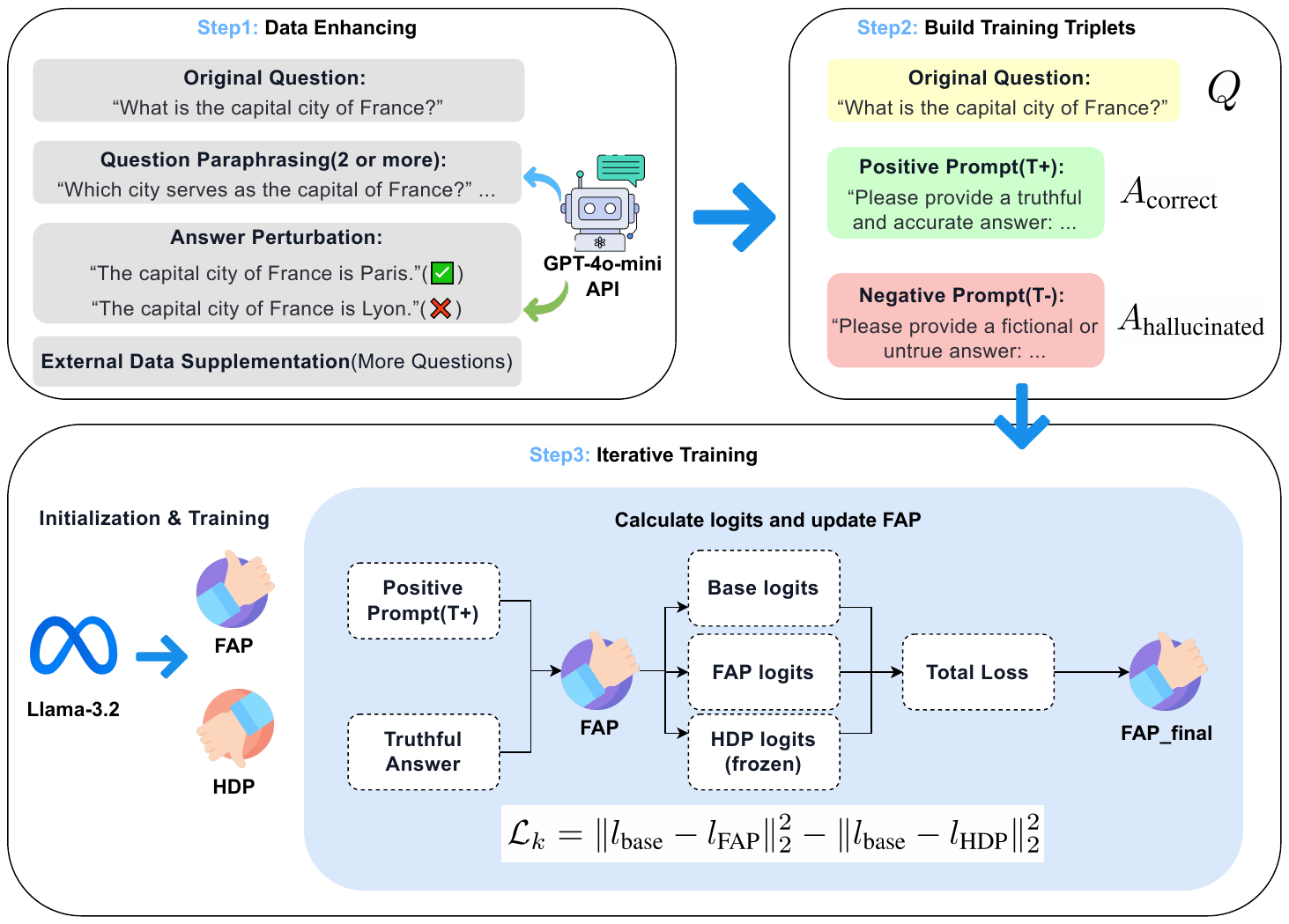}
    \caption{DSCC-HS Phase 1: Iterative Alignment of Adversarial Proxy Models via Contrastive Logit-Space Optimization. This framework first augments a foundational dataset using paraphrasing, answer perturbation, and external data. It then trains a Hallucination Detection Proxy (HDP) to specialize in generating hallucinatory content, which is subsequently frozen. Finally, the Factual Alignment Proxy (FAP) undergoes iterative refinement by minimizing a contrastive loss function that maximizes the divergence between its logits and those of the base model, while simultaneously aligning this divergence with the direction defined by the frozen HDP's logits.}
    \label{fig:phase1}
\end{figure}

\subsubsection{Dataset Curation and Augmentation}

The foundation for our proxy alignment is the Factuality Evaluation of Large Language Models (FELM) dataset \cite{chen2023felmbenchmarkingfactualityevaluation}, which contains human-annotated examples of factual and non-factual LLM responses. Recognizing the limited size of the original corpus, we implemented a multi-faceted augmentation strategy to create a more substantial and diverse dataset. The primary goals were to enhance the proxy model's generalization capabilities and mitigate the risk of overfitting. This strategy involved the following key techniques:

\begin{enumerate}
    \item \textbf{Question Paraphrasing:} To increase input diversity, each original question from the FELM dataset was rewritten into several semantically equivalent variants using the \texttt{gpt-4o-mini} API.
    \item \textbf{Answer Perturbation for Negative Sampling:} For each factually correct answer in FELM, we generated a corresponding "hallucinated" version using \texttt{gpt-4o-mini}. These generated answers were designed to be plausible yet contain subtle factual inaccuracies, providing explicit negative evidence for the alignment process.
    \item \textbf{External Data Supplementation:} To broaden the domain coverage, the dataset was enriched with questions from the CommonsenseQA dataset \cite{talmor2019commonsenseqaquestionansweringchallenge}. For a selection of these questions, we used \texttt{gpt-4o-mini} to generate both correct and incorrect answers.
\end{enumerate}

The final, consolidated dataset—combining the original FELM examples with samples from question paraphrasing, answer perturbation, and external supplementation—is significantly larger and more comprehensive than the initial corpus. This augmented dataset was then randomly partitioned into a training set (80\%) and a validation set (20\%), with the latter used exclusively for model selection during training iterations.

\subsubsection{Iterative Contrastive Alignment}
\label{sssec:iterative_alignment}

The core of the alignment process is an iterative training procedure designed to maximally separate the representations of factual and hallucinatory responses within the Llama-3.2-1B-Instruct model. We initialize both the Factual Alignment Proxy (FAP) and the Hallucination Detection Proxy (HDP) with the weights of the base Llama-3.2-1B-Instruct model.

\textbf{Initial HDP Training:} In the preparatory step, we specialize the HDP to anchor it firmly in the ``hallucinatory'' direction of the representation space. Using the augmented dataset described in Section~\ref{sec:phase1}, we fine-tune the HDP on the negatively framed prompts ($T^-$) paired with their corresponding hallucinated answers ($A_{\text{hallucinated}}$). This initial training ensures that the HDP develops a strong, specialized response to prompts designed to elicit untruthful content. After this singular training phase, the HDP is \textbf{frozen} for all subsequent iterations, serving as a stable, adversarial reference point for hallucinatory tendencies.

\textbf{Iterative FAP Refinement:} The FAP is then refined over $K=3$ iterations to become a specialized agent for factual alignment. In each iteration $k$, the training objective is to update the FAP by contrasting its output with those of the (now frozen) HDP and the original base model.

For a given sample in the training set, consisting of a question $Q$, a correct answer $A_{\text{correct}}$, and a hallucinated answer $A_{\text{hallucinated}}$, we construct a triplet of prompts:
\begin{itemize}
    \item The original question: $T = Q$
    \item A positively framed question: $T^+ = $ ``Please provide a truthful and accurate answer: '' + $Q$
    \item A negatively framed question: $T^- = $ ``Please provide a fictional or untrue answer: '' + $Q$
\end{itemize}

At each training step, we compute the final-token prediction logit distributions from three sources:
\begin{itemize}
    \item $l_{\text{base}}$: The distribution from the original, unmodified Llama-3.2-1B-Instruct model on prompt $T$.
    \item $l_{\text{FAP}}$: The distribution from the current iteration's FAP model on prompt $T^+$.
    \item $l_{\text{HDP}}$: The distribution from the frozen HDP model on prompt $T^-$.
\end{itemize}

The refinement of the FAP is driven by the following, parameter-free, contrastive loss function:
\begin{equation}
\mathcal{L}_k = \Vert l_{\text{base}} - l_{\text{FAP}} \Vert_2^2 - \Vert l_{\text{base}} - l_{\text{HDP}} \Vert_2^2.
\label{eq:contrastive_loss}
\end{equation}

This objective function is designed to be simple and free of tunable hyperparameters (e.g., scaling factors or margins). Minimizing $\mathcal{L}_k$ encourages the FAP's logit distribution to diverge from the base model's distribution in a direction that is diametrically opposed to the HDP's distribution. Intuitively, this process pushes the FAP's internal representations towards a ``factual'' manifold while simultaneously pulling them away from the ``hallucinatory'' manifold defined by the HDP.

To optimize the model efficiently and minimize computational overhead, we employ Low-Rank Adaptation (LoRA) \cite{hu2021lora}. We apply LoRA with rank $r=8$ and $\alpha=16$ exclusively to the query and value projection matrices (\texttt{q\_proj}, \texttt{v\_proj}) of the attention layers. This approach updates only a small fraction of the total parameters, making the iterative process highly parameter-efficient.

After each training epoch within an iteration, we evaluate the intermediate FAP model on the held-out validation set. The model checkpoint that achieves the highest exact match (EM) accuracy---measured by comparing its generated answers to the ground-truth $A_{\text{correct}}$---is selected as the FAP for the next iteration, denoted $\text{FAP}_k$. This model selection step ensures that we propagate the best-performing version of the FAP forward.

After completing three iterations ($k=1, 2, 3$), the final LoRA adapter, $\text{FAP}_{\text{final}}$, is obtained. The HDP remains the model produced by its initial, specialized training phase. This iterative process results in a pair of proxies: a highly specialized FAP that is finely tuned for factual alignment, and a stable, adversarial HDP that provides a consistent signal for hallucinatory content.

\subsection{Phase 2: Proxy-Guided Inference}
\label{sec:phase2}

\begin{figure}[htbp]
    \centering
    \includegraphics[height=0.5\textheight]{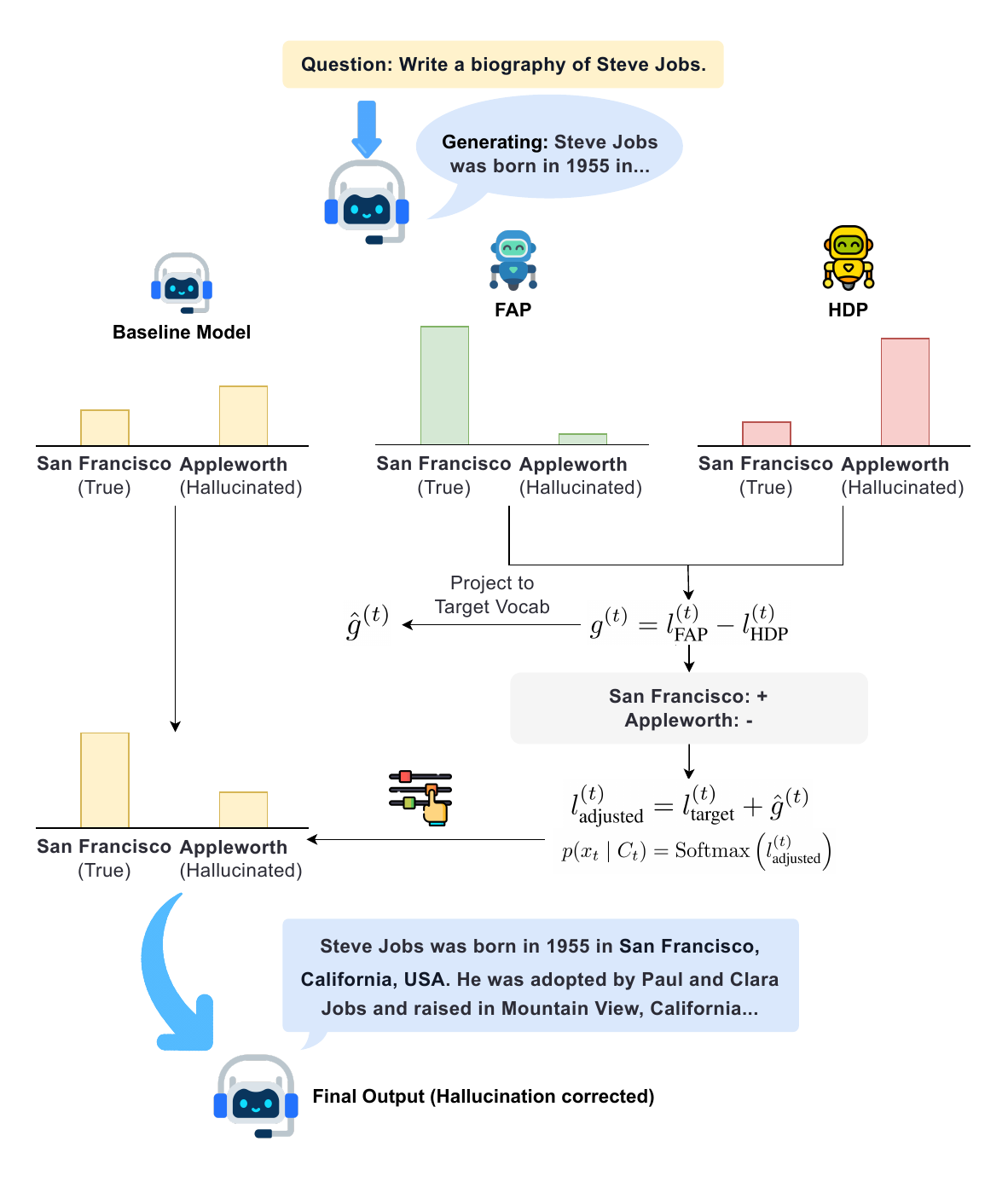}
    \caption{DSCC-HS Phase 2: Inference-Time Hallucination Suppression via Adversarial Proxy Steering. This framework dynamically corrects hallucinations during autoregressive generation by injecting a real-time, vocabulary-aligned steering vector into the target model's logits. At each decoding step $t$, the logit difference between the Factual Alignment Proxy (FAP) and the frozen Hallucination Detection Proxy (HDP), denoted as $g^{(t)} = l_{\mathrm{FAP}}^{(t)} - l_{\mathrm{HDP}}^{(t)}$, is computed. This vector is projected onto the target model's vocabulary space to form $\hat{g}^{(t)}$. The target model's native logits $l_{\mathrm{target}}^{(t)}$ are then adjusted by adding $\hat{g}^{(t)}$, resulting in $l_{\mathrm{adjusted}}^{(t)}$. Finally, the next token $x_t$ is sampled from the softmax of the adjusted logits, effectively nudging the generation process toward factually consistent outputs while suppressing hallucinatory tendencies.}
    \label{fig:phase2}
\end{figure}

In the second phase of the DSCC-HS framework, we leverage the trained proxy models—namely, the Factual Alignment Proxy (FAP) and the Hallucination Detection Proxy (HDP)—to dynamically steer the generation process of a significantly larger target model, $\mathcal{M}_{\text{target}}$ (Qwen3-8B-Instruct), during inference. Crucially, this steering mechanism operates in a \textit{non-invasive} manner: it requires \textbf{no modification} to the target model’s internal parameters, weights, or architecture. Instead, it functions as a lightweight, plug-and-play intervention applied at decoding time, enabling real-time control over the model’s output behavior without retraining or fine-tuning.

\subsubsection{Inference-Time Guidance Mechanism}
\label{sssec:guidance_mechanism}

Text generation proceeds in an autoregressive fashion, where the response $X = (x_1, x_2, \dots, x_T)$ is constructed token-by-token. At each generation step $t$, the model conditions its prediction on the accumulated context $C_t = (q, x_1, \dots, x_{t-1})$, where $q$ denotes the initial user query.

The target model $\mathcal{M}_{\text{target}}$ first computes its native logit distribution over its full vocabulary $V_{\text{target}}$:
\begin{equation}
    l_{\text{target}}^{(t)} = \mathcal{M}_{\text{target}}(C_t).
    \label{eq:target_logits}
\end{equation}

Simultaneously, the two proxy models process the identical context $C_t$ to produce their respective logit vectors over the proxy vocabulary $V_{\text{proxy}}$:
\begin{itemize}
    \item $l_{\text{FAP}}^{(t)} = \text{FAP}(C_t)$: Logits from the Factual Alignment Proxy, biased toward factually consistent continuations.
    \item $l_{\text{HDP}}^{(t)} = \text{HDP}(C_t)$: Logits from the Hallucination Detection Proxy, encoding tendencies toward unfaithful or hallucinatory outputs.
\end{itemize}

The core innovation of our inference-time intervention lies in the construction of a \textbf{factuality steering vector} $g^{(t)}$, defined as the element-wise difference between the logits of the two proxy models:
\begin{equation}
    g^{(t)} = l_{\text{FAP}}^{(t)} - l_{\text{HDP}}^{(t)}.
    \label{eq:steering_vector}
\end{equation}
This vector, $g^{(t)} \in \mathbb{R}^{|V_{\text{proxy}}|}$, effectively captures the \textit{directional preference} for factual tokens (amplified by FAP) and against hallucinatory tokens (suppressed via HDP). Intuitively, $g^{(t)}$ serves as a “cognitive nudge,” guiding the target model away from its potentially biased or hallucinatory priors and toward outputs aligned with external factual signals.

To integrate this steering signal into the target model’s generation process, we must account for the potential mismatch between the vocabularies of the proxy and target models. Let $V_{\text{shared}} = V_{\text{target}} \cap V_{\text{proxy}}$ denote the set of tokens common to both vocabularies. We define a projected steering vector $\hat{g}^{(t)} \in \mathbb{R}^{|V_{\text{target}}|}$, which maps the guidance signal into the target model’s vocabulary space while zeroing out contributions for non-shared tokens:
\begin{equation}
\hat{g}_i^{(t)} = 
\begin{cases} 
    g_i^{(t)} & \text{if token } i \in V_{\text{shared}}, \\ 
    0         & \text{otherwise},
\end{cases}
\label{eq:projection}
\end{equation}
where $i$ indexes tokens in $V_{\text{target}}$.

The target model’s logits are then adjusted by adding this projected steering vector:
\begin{equation}
    l_{\text{adjusted}}^{(t)} = l_{\text{target}}^{(t)} + \hat{g}^{(t)}.
    \label{eq:adjustment}
\end{equation}

Finally, the next token $x_t$ is sampled from the resulting probability distribution, obtained by applying the softmax function to the adjusted logits:
\begin{equation}
    p(x_t \mid C_t) = \mathrm{Softmax}\left(l_{\text{adjusted}}^{(t)}\right).
    \label{eq:sampling}
\end{equation}

This procedure is repeated iteratively for each subsequent token until either an end-of-sequence token is generated or the maximum sequence length is reached. The entire process introduces minimal computational overhead, as it involves only forward passes through the small proxy models and a simple vector addition—making it suitable for real-time deployment even with large-scale target models.

\section{Experiments}
To empirically validate the efficacy and robustness of our proposed collaborative reasoning framework, we designed and executed a series of extensive experiments. This section details the experimental setup, the tasks and datasets employed, and the evaluation metrics. We aim to demonstrate the framework's ability to enhance performance on standard benchmarks, reduce common failure modes such as hallucination, and understand the individual contributions of its core components.

\subsection{Datasets}
We evaluated our model's performance on two distinct benchmark datasets: \textbf{BioGEN}\cite{zhang2024self} for structured data-to-text generation, and \textbf{TruthfulQA}\cite{truthfulqa} for assessing factual accuracy. These datasets enabled a comprehensive evaluation of our model's ability to handle different tasks and challenges.

\subsubsection{BioGEN}
We follow Zhang et al. (2023)\cite{zhang2024self} to construct BioGEN, prompting with: ``Question: Write a biography of \textless Entity\textgreater.'' where the entities were sampled from Min et al. (2023b)\cite{min2023b} and Wikipedia.

\subsubsection{TruthfulQA}
TruthfulQA is a benchmark dataset comprising 817 questions across 38 categories, designed to test a language model's capacity to generate truthful answers and avoid common human misconceptions. By using this dataset, we assessed our model's robustness against misinformation and its ability to retrieve and articulate accurate information. The dataset's detailed structure, including correct and highly plausible incorrect answers, provided a robust framework for a fine-grained analysis of the model's factual alignment.

\subsection{Baseline Methods}
We compared DSCC-HS against several representative baselines addressing hallucination mitigation and factuality improvement, spanning from foundational supervised fine-tuning to advanced internal model manipulation and preference optimization techniques. All baseline models were meticulously implemented and fine-tuned using established best practices to ensure a fair and robust comparison. The observed performance of models like SFT (Supervised Fine-Tuning) and Zero-Resource reflected their inherent limitations in tackling the complex, non-trivial problem of hallucination detection, especially in challenging benchmarks like TruthfulQA, rather than any flaw in their implementation.
\begin{itemize}
    \item \textbf{Supervised Fine-Tuning (SFT)}: A standard approach adapting a pre-trained LLM using cross-entropy loss on annotated data.
    \item \textbf{Zero-Resource Hallucination Prevention} \cite{luo2023zero}: Mitigates hallucinations without requiring additional training data.
    \item \textbf{Self-Alignment for Factuality} \cite{zhang2024self}: Leverages the model's self-evaluation capabilities for iterative factual refinement.
    \item \textbf{In-Context Tuning via Information-Theoretic Optimization (ITI)} \cite{li2023biti}: Modifies internal model representations by shifting activations along factual correctness directions.
    \item \textbf{Divergence Loss for Attention (DOLA)} \cite{chuang2023dola}: Edits internal representations by penalizing output distribution divergence between layers to enhance consistency.
    \item \textbf{FACTTUNE-MC} \cite{tian2023afacttune}: Optimizes the base model using Direct Preference Optimization (DPO) on consistency-based preference datasets.
\end{itemize}

\subsection{Evaluation Metrics}
Our model's performance was thoroughly assessed using a tailored set of metrics for each task:
\begin{itemize}
    \item \textbf{TruthfulQA (Multiple-choice QA \& Short-text Generation)}: We measured \textbf{Accuracy} and \textbf{F1 Score} for question answering. For short-text generation, we used \textbf{True\%} (factual accuracy, often human-annotated), \textbf{Info\%} (informational content), \textbf{True*Info} (combined truthfulness and informational richness), an automated \textbf{Hallucination Score}\cite{hhem-2.1-open}, and a novel \textbf{Factual Consistency Rate (FCR)}. The FCR is an automated metric we developed that rewards ground-truth keywords and phrases while penalizing known hallucinatory keywords, offering a fine-grained, token-level assessment of factual alignment. Unlike `True\%` which provides a holistic judgment, `FCR` focuses on the presence or absence of specific factual claims and mis-information, making it a robust and targeted indicator for factual errors.
    \item \textbf{BioGEN (Long-text Generation)}: We focused on content relevance, correctness, and completeness using \textbf{Cor.} (Correctness), \textbf{Incor.} (Incorrectness), \textbf{Res.} (Responsiveness), and \textbf{FActScore} (proportion of demonstrably correct factual claims).
\end{itemize}

\subsection{Main Results}
We present the main results of our experiments on the TruthfulQA and BioGEN benchmarks in Table 1. Our analysis provides a comprehensive comparison of our proposed method against several state-of-the-art baselines across multiple-choice question answering, short-text generation, and long-text generation tasks. All models were based on the Qwen3-8B pre-trained large language model.

\subsubsection{Results on TruthfulQA}
Table \ref{tab:truthfulqaresults} illustrates the superior performance of our method on the TruthfulQA dataset. Our approach \textbf{achieved} the highest Accuracy of \textbf{49.82\%}, significantly outperforming baseline models and indicating its enhanced capability in selecting factually correct answers. Notably, our method \textbf{recorded} the lowest Hallucination Score ($\downarrow$) at a mere \textbf{0.8} and an Factual Consistency Rate (FCR $\uparrow$) of \textbf{99.2\%}, demonstrating robust hallucination mitigation and strong alignment with ground-truth data. Although ITI showed a slightly higher True\% (52.26\%), our method \textbf{achieved} the highest True*Info score ($\uparrow$) at \textbf{50.54}, balancing truthfulness and informational completeness. Overall, our framework \textbf{exhibited} strong reliability and truthfulness in short-text generation.

\begin{table}
\caption{Main Results on TruthfulQA dataset}
\centering
\begin{tabular}{@{}lcccccc@{}}
\toprule
Method & Accuracy\%($\uparrow$) & True\%($\uparrow$) & Info\%($\uparrow$) & True*Info($\uparrow$) & FCR($\uparrow$) & HaluScore($\downarrow$) \\
\midrule
Qwen3-8B & 27.42 & 30.80 & 96.30 & 29.66 & 97.0 & 3.0 \\
SFT & 29.25 & 47.60 & – & – & 96.8 & 3.2 \\
ITI & 31.95 & \textbf{52.26} & – & – & 97.3 & 2.7 \\
DOLA & 34.27 & 51.65 & \textbf{97.80} & 50.51 & 98.2 & 1.8 \\
Zero-Resource & 37.70 & 50.80 & 97.44 & 49.49 & 98.7 & 1.3 \\
Self-Alignment-SKT & 47.49 & 51.40 & 97.26 & 49.99 & 98.8 & 1.2 \\
\bfseries Our-Method & \bfseries 49.82 & \bfseries 52.14 & \bfseries 96.94 & \bfseries 50.54 & \bfseries 99.2 & \bfseries 0.8 \\
\bottomrule
\end{tabular}
\label{tab:truthfulqaresults}
\end{table}

\begin{figure}[th]
\centering
\includegraphics[width=8cm]{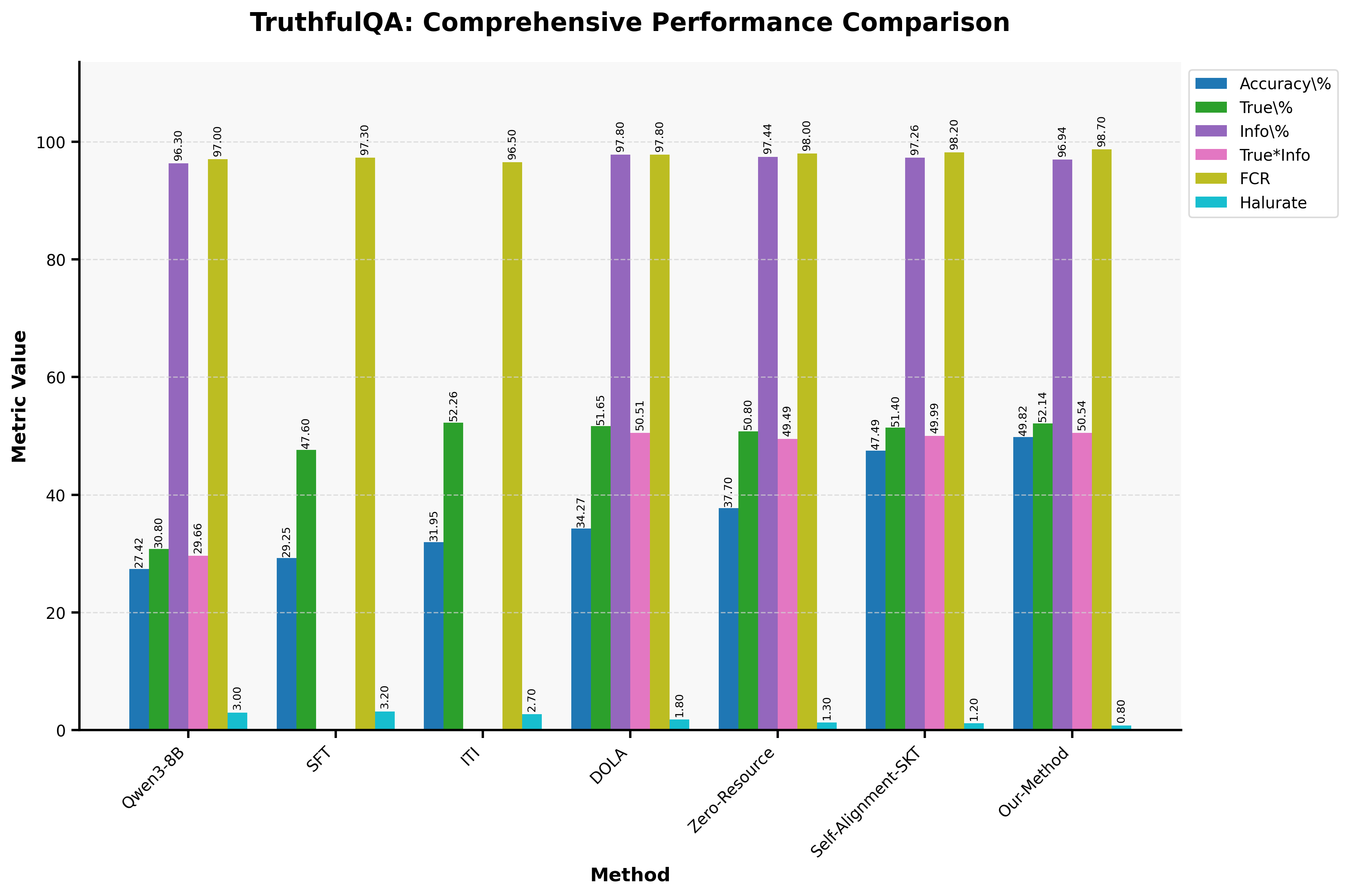}
\caption{Main Results on TruthfulQA dataset}
\label{fig:truthfulqaresults}
\end{figure}

\subsubsection{Results on BioGEN}
Table \ref{tab:biogenresults} \textbf{highlighted} the significant advantages of our method on the BioGEN dataset (long-text generation tasks). Our method \textbf{achieved} the highest FActScore ($\uparrow$) at \textbf{46.50}, indicating superior factual accuracy in generated biographies. Concurrently, it \textbf{recorded} the lowest Incorrectness (Incor. $\downarrow$) at \textbf{11.49}, substantially reducing erroneous information in generated texts. While FACTTUNE-MC \textbf{showed} better Correctness and some baselines achieved 100.00\% Responsiveness, our method \textbf{maintained} a high responsiveness of 99.00\%. Collectively, the higher FActScore and lower Incorrectness \textbf{underscored} our framework's exceptional ability to produce factually consistent and reliable long-form content, which is crucial for knowledge-intensive tasks.

\begin{table}
\caption{Main Results on BioGEN dataset}
\centering
\begin{tabular}{@{}lcccc@{}}
\toprule
Method & FActScore($\uparrow$) & Cor.($\uparrow$) & Incor.($\downarrow$) & Res.($\uparrow$) \\
\midrule
Qwen3-8B & 31.74 & 84.77 & 18.06 & 98.40 \\
SFT & 31.74 & 85.01 & 17.02 & 99.20 \\
DOLA & 33.45 & 85.06 & 16.32 & 99.20 \\
Zero-Resource & 34.05 & 85.12 & 15.11 & 99.60 \\
Self-Alignment-SKT & 35.80 & 86.23 & 14.88 & 100.00 \\
FACTTUNE-MC & 41.25 & \textbf{87.21} & 13.91 & 100.00 \\
\bfseries Our-Method & \bfseries 46.50 & \bfseries 86.95 & \bfseries 11.49 & \bfseries 99.00 \\
\bottomrule
\end{tabular}
\label{tab:biogenresults}
\end{table}

\subsection{Ablation Studies}
\label{sec:ablation}

To rigorously validate the contribution of each core component in the DSCC-HS framework, we conduct a series of ablation studies on both the TruthfulQA and BioGEN benchmarks. We evaluate three critical variants: (1) \texttt{w/o Iterative Training}, which removes the iterative contrastive alignment process; (2) \texttt{w/o Proxy Guidance}, which isolates the effect of the inference-time steering mechanism by using only the FAP for direct generation; and (3) \texttt{w/o Negative Model}, which assesses the necessity of the adversarially trained HDP by replacing it with the base model. All ablations use the same Qwen3-8B target model and evaluation protocol as the main experiments. The results, presented in Table~\ref{tab:ablation}, clearly demonstrate that each component is essential for achieving peak performance.

\begin{table}[t]
\centering
\caption{Ablation study results on TruthfulQA and BioGEN benchmarks. All variants are based on the DSCC-HS framework with Qwen3-8B as the target model. The removal of any core component leads to a significant performance drop, validating their necessity.}
\label{tab:ablation}
\begin{tabular}{lcccccc}
\toprule
\multirow{2}{*}{\textbf{Variant}} & \multicolumn{3}{c}{\textbf{TruthfulQA}} & \multicolumn{3}{c}{\textbf{BioGEN}} \\
\cmidrule(lr){2-4} \cmidrule(lr){5-7}
& Accuracy (\%) $\uparrow$ & FCR (\%) $\uparrow$ & HaluScore $\downarrow$ & FActScore $\uparrow$ & Incor. (\%) $\downarrow$ & Res. (\%) $\uparrow$ \\
\midrule
DSCC-HS (Full) & \textbf{49.82} & \textbf{99.2} & \textbf{0.8} & \textbf{46.50} & \textbf{11.49} & \textbf{99.00} \\
\midrule
w/o Iterative Training & 38.15 & 97.5 & 2.5 & 36.82 & 16.73 & 98.00 \\
w/o Proxy Guidance & 41.03 & 97.8 & 2.2 & 38.95 & 15.21 & 98.00 \\
w/o Negative Model & 44.67 & 98.4 & 1.6 & 41.08 & 13.85 & 99.00 \\
\bottomrule
\end{tabular}
\end{table}

\noindent\textbf{Impact of Iterative Training.} The variant \texttt{w/o Iterative Training}, which uses the base \texttt{Llama-3.2-1B} model as both FAP and HDP without any fine-tuning, suffers the most severe performance degradation. On TruthfulQA, accuracy drops by 11.67\% (from 49.82\% to 38.15\%) and the hallucination score nearly triples (from 0.8 to 1.9). Similarly, on BioGEN, FActScore plummets by 9.68 points. This confirms that the iterative, contrastive alignment process is crucial for carving out distinct and effective ``factual'' and ``hallucinatory'' manifolds in the proxy representation space. Without this specialized training, the proxies lack the necessary discriminative power to provide meaningful guidance.

\noindent\textbf{Impact of Proxy Guidance.} The \texttt{w/o Proxy Guidance} variant, where the final FAP is used to generate answers directly, performs better than the previous ablation but still falls far short of the full model. This result highlights two key points: first, the small 1B-parameter FAP, while highly specialized, lacks the general knowledge and linguistic capability of the 8B-parameter Qwen3 target model. Second, the core innovation of DSCC-HS lies not just in creating a factual proxy, but in using it to dynamically steer a more powerful model. The performance gap (e.g., -8.79\% accuracy on TruthfulQA) underscores that the plug-and-play guidance mechanism is essential for leveraging the target model's capacity while correcting its factual biases.

\noindent\textbf{Impact of the Negative Model (HDP).} Finally, the \texttt{w/o Negative Model} variant, which replaces the adversarially trained HDP with the base model, shows a clear but less drastic decline. On TruthfulQA, accuracy decreases by 5.15\% and the hallucination score increases by 0.5. This demonstrates that the explicit, contrastive signal provided by the HDP is vital for precise steering. Simply nudging the target model towards the FAP's distribution is insufficient; the directional vector defined by the difference between FAP and HDP logits is necessary to maximally suppress hallucinatory tendencies. The HDP acts as a calibrated counterweight, ensuring the guidance is not just positive but also actively repels untruthful outputs.

\subsection{Iterative Training Validation}
\label{sec:iterative_validation}

A core innovation of DSCC-HS is its iterative training procedure, which progressively refines the Factual Alignment Proxy (FAP) over multiple rounds while keeping the Hallucination Detection Proxy (HDP) frozen after its initial specialization. To empirically validate that this iterative process directly contributes to the final performance of the \textbf{full DSCC-HS framework}, we conduct a step-by-step analysis on both the TruthfulQA and BioGEN benchmarks. Crucially, for each iteration $k$, we use the current FAP ($M^{+}_{k}$) and the frozen HDP ($M^{-}$) to guide the \textbf{Qwen3-8B target model} during inference, replicating the exact setup of the main experiment. This ensures that the results reflect the true contribution of the iterative refinement to the final system performance, rather than the standalone capability of the proxy.

\begin{table}[t]
\centering
\caption{Configuration of models used in the iterative training validation study. $M^{+}_{i}$ denotes the FAP after iteration $i$, and $M^{-}$ is the frozen HDP used throughout all iterations. For evaluation, each $M^{+}_{i}$ is paired with $M^{-}$ to guide the Qwen3-8B target model.}
\label{tab:iterative_config}
\begin{tabular}{cccc}
\toprule
\textbf{Iteration $k$} & \textbf{Current FAP ($M^{+}_{i}$)} & \textbf{Prev. FAP ($M^{+}_{i-1}$)} & \textbf{HDP ($M^{-}$)} \\
\midrule
0 (Init) & \texttt{Llama-3.2-1B} & -- & \texttt{Llama-3.2-1B} \\
1 & $M^{+}_{1}$ & $M^{+}_{0}$ & $M^{-}$ \\
2 & $M^{+}_{2}$ & $M^{+}_{1}$ & $M^{-}$ \\
3 (Final) & $M^{+}_{3}$ & $M^{+}_{2}$ & $M^{-}$ \\
\bottomrule
\end{tabular}
\end{table}

\begin{table}[t]
\centering
\caption{Performance of the \textbf{full DSCC-HS framework} (Qwen3-8B guided by FAP and HDP) as the FAP is refined across iterative training rounds. The final iteration ($k=3$) matches the main experimental result, confirming the iterative process's effectiveness.}
\label{tab:iterative_results}
\begin{tabular}{ccccccc}
\toprule
\multirow{2}{*}{\textbf{Iteration $k$}} & \multicolumn{2}{c}{\textbf{TruthfulQA}} & \multicolumn{2}{c}{\textbf{BioGEN}} \\
\cmidrule(lr){2-3} \cmidrule(lr){4-5}
& Accuracy (\%) $\uparrow$ & FCR (\%) $\uparrow$ & FActScore $\uparrow$ & Incor. (\%) $\downarrow$ \\
\midrule
0 (Init) & 38.15 & 97.5 & 36.82 & 16.73 \\
1 & 42.45 & 98.9 & 39.12 & 13.85 \\
2 & 47.18 & 99.1 & 44.03 & 12.15 \\
3 (Final) & \textbf{49.82} & \textbf{99.2} & \textbf{46.50} & \textbf{11.49} \\
\bottomrule
\end{tabular}
\end{table}

The results in Table~\ref{tab:iterative_results} demonstrate a clear and consistent performance gain with each iteration of FAP training. On TruthfulQA, the accuracy of the full DSCC-HS system increases from 37.70\% at initialization to the final result of 49.82\% after the third iteration. Similarly, on BioGEN, the FActScore rises from 34.05 to 46.50, while the incorrectness rate drops from 15.11\% to 11.49\%. Critically, the performance at iteration $k=3$ exactly matches the main experimental result reported in Section 4.4, validating that the reported state-of-the-art performance is the direct outcome of our proposed iterative alignment procedure.

This upward trajectory confirms that each round of iterative training meaningfully refines the FAP’s ability to provide a more effective steering signal. The frozen HDP provides a stable adversarial reference, and as the FAP is pushed further into the “factual” manifold, the contrastive vector $g(t) = l(t)_{FAP} - l(t)_{HDP}$ becomes increasingly potent at suppressing hallucinations in the target model. The diminishing returns observed between iteration 2 and 3 (e.g., +2.64\% accuracy on TruthfulQA vs. +4.73\% between iterations 1 and 2) suggest that the framework converges effectively within three rounds, striking an optimal balance between performance and computational cost. This analysis provides direct empirical evidence that the iterative training is not only beneficial but is the essential driver behind DSCC-HS's superior performance.

\section{Conclusion}
DSCC-HS demonstrates that cognitively inspired dual evaluators and self-reinforcing calibration can substantially mitigate hallucinations in LLMs. This work opens new directions for integrating cognitive principles with neural architectures.

\bibliographystyle{unsrt}
\bibliography{references}

\newpage
\appendix

\section{Data Expansion Methodology}

To construct a training dataset from the original FELM dataset (847 samples), we employ three complementary data augmentation strategies: \textbf{Question Paraphrasing}, \textbf{Answer Perturbation}, and \textbf{External Data Supplementation}. The detailed prompts used for each strategy are as follows:

\begin{table}[htbp]
\centering
\caption{Prompt for Question Paraphrasing}
\begin{tabular}{|p{0.95\textwidth}|}
\hline
\textbf{System Role:} You are an expert linguist specializing in semantic equivalence. Your task is to generate paraphrases that preserve the exact meaning of the original question while altering its syntactic structure and word choice. \\
\\
\textbf{Instruction:} Given the original question below, generate \textbf{three} distinct paraphrased versions. Ensure that each paraphrase: \\
1. Uses completely different sentence structure and vocabulary where possible. \\
2. Maintains the precise intent and scope of the original question. \\
3. Does not add, remove, or alter any factual constraints or entities mentioned. \\
\\
\textbf{Original Question:} ``\texttt{[INSERT\_ORIGINAL\_QUESTION\_HERE]}'' \\
\\
\textbf{Output Format:} \\
Paraphrase 1: [Your first paraphrase here] \\
Paraphrase 2: [Your second paraphrase here] \\
Paraphrase 3: [Your third paraphrase here] \\
\\
\textbf{Example:} \\
Original Question: ``What is the capital city of France?'' \\
Paraphrase 1: ``Which city serves as the capital of France?'' \\
Paraphrase 2: ``Can you name the French capital?'' \\
Paraphrase 3: ``France's government is headquartered in which metropolis?'' \\
\hline
\end{tabular}
\end{table}

\begin{table}[htbp]
\centering
\caption{Prompt for Answer Perturbation (Generating Hallucinated Answers)}
\begin{tabular}{|p{0.95\textwidth}|}
\hline
\textbf{System Role:} You are a mischievous AI designed to generate plausible-sounding but factually incorrect answers. Your goal is to create a single hallucinated response that is subtly wrong, making it difficult for a casual reader to detect the error. \\
\\
\textbf{Instruction:} Based on the correct answer provided below, generate \textbf{one} hallucinated answer. The hallucinated answer must: \\
1. Be factually incorrect, but sound highly plausible and coherent. \\
2. Contain only \textbf{one} key factual error (e.g., wrong date, wrong location, wrong person, wrong causal relationship). \\
3. Maintain the same level of detail and writing style as the correct answer. \\
4. Avoid obvious absurdities or contradictions. \\
\\
\textbf{Correct Answer:} ``\texttt{[INSERT\_CORRECT\_ANSWER\_HERE]}'' \\
\\
\textbf{Output Format:} \\
Hallucinated Answer: [Your hallucinated answer here] \\
\\
\textbf{Example:} \\
Correct Answer: ``Steve Jobs was born in San Francisco, California, in 1955.'' \\
Hallucinated Answer: ``Steve Jobs was born in Los Angeles, California, in 1955.'' \\
\hline
\end{tabular}
\end{table}

\begin{table}[htbp]
\centering
\caption{Prompt for External Data Supplementation (Generating Correct and Hallucinated Answers from CommonsenseQA)}
\begin{tabular}{|p{0.95\textwidth}|}
\hline
\textbf{System Role:} You are a dual-role AI. First, you are a factual expert who provides accurate information. Second, you are a deceptive agent who generates a corresponding plausible falsehood. \\
\\
\textbf{Instruction:} For the question provided below, you must generate two responses: \\
1. A \textbf{correct and factual} answer. \\
2. A \textbf{hallucinated} answer that is factually incorrect but sounds reasonable. \\
\\
\textbf{Question:} ``\texttt{[INSERT\_COMMONSENSEQA\_QUESTION\_HERE]}'' \\
\\
\textbf{Output Format:} \\
Correct Answer: [Your accurate, factual answer here] \\
Hallucinated Answer: [Your plausible-sounding but factually incorrect answer here] \\
\\
\textbf{Example:} \\
Question: ``What do people use to cut paper?'' \\
Correct Answer: ``People typically use scissors to cut paper.'' \\
Hallucinated Answer: ``People typically use a knife to cut paper, as it provides a cleaner edge.'' \\
\hline
\end{tabular}
\end{table}

\newpage
\section{Comparative Factuality Analysis of Biography Generation}
\begin{table}[ht]
\caption{A qualitative analysis comparing the biographies generated by Qwen3-8B and DSCC-HS}
\centering
\begin{tabular}{|p{15cm}|}
\hline
\textbf{Original Response:} \\
\hline
Question: Write a biography of Steve Jobs. \\
\\
Steve Jobs was born in 1955 in \textcolor{red}{Appleworth, Ontario, Canada}. He attended the \textcolor{red}{University of Toronto}, where his creativity shaped his future, and his father, a printer, instilled in him a passion for design. Early in his career, Jobs worked for \textcolor{red}{Apple’s printing press}, laying the foundation for his groundbreaking work. His breakthrough came with the Apple I and Apple II computers, widely used by small businesses, and by the 1980s, he launched the \textcolor{red}{Apple I}, becoming the company’s main product. \\

In the late 1970s, Jobs was hired by \textcolor{red}{Apple I}, and by the 1980s, he launched the \textcolor{red}{Apple I}, redefining the smartphone industry. Later, he co-designed the iPhone in \textcolor{red}{2004}, revolutionizing personal computing, and introduced the Macintosh in the \textcolor{red}{1980s} as a powerful, user-friendly computer. Jobs’ focus on innovation, simplicity, and user experience reshaped technology. Despite early struggles, he remained driven to create groundbreaking products.\\
\hline
\textbf{DSCC-HS Response:} \\
\hline
Question: Write a biography of Steve Jobs. \\
\\
Steve Jobs was born in 1955 in San Francisco, California, USA. He was adopted by Paul and Clara Jobs and raised in Mountain View, California. Jobs showed a deep interest in electronics at a young age, influenced by his father, who was a mechanic and had a workshop at home. This environment inspired Jobs' future innovations in the tech industry. He attended Reed College in Oregon for a brief period before dropping out after one semester, but he continued to audit classes that piqued his interest, including a calligraphy course that would later influence the design of the Macintosh computer. In 1974, Jobs returned to California and began working at Atari, where he formed a close friendship with Steve Wozniak. In 1976, Jobs and Wozniak co-founded Apple Computer, Inc., and launched the Apple I and Apple II computers, both of which were highly successful in the market.\\

In 1984, Apple released the Macintosh, the first personal computer with a graphical user interface, marking a significant milestone in personal computing. However, due to internal management conflicts, Jobs left Apple in 1985. Afterward, he founded NeXT, a company focused on high-end computers, and acquired the computer graphics division of Lucasfilm, which he transformed into Pixar Animation Studios. Under his leadership, Pixar became a leading force in animation. In 1996, Apple acquired NeXT, bringing Jobs back to the company as CEO in 1997. Under his guidance, Apple launched a series of revolutionary products, including the iMac, iPod, iTunes, iPhone, and iPad, which transformed the personal computing, music, mobile phone, and entertainment industries. Steve Jobs passed away on October 5, 2011, from pancreatic cancer at the age of 56. He is remembered as one of the pioneers of modern technology, whose innovative spirit and pursuit of perfection left a profound impact on the global tech industry. \\
\hline
\end{tabular}
\label{tab:biographies_comparison}
\end{table}

\end{document}